\pdfoutput=1

\documentclass[11pt]{article}

\usepackage{acl}

\usepackage{times}
\usepackage{latexsym}

\usepackage[T1]{fontenc}

\usepackage[utf8]{inputenc}

\usepackage{microtype}

\usepackage{inconsolata}

\usepackage{graphicx}
\usepackage{hyperref}
\usepackage{booktabs}
\usepackage{multirow}
\usepackage{placeins}
\usepackage{tabularx}

\title{TUM-MiKaNi at SemEval-2025 Task 3: \textit{Our Title}}
\title{TUM-MiKaNi at SemEval-2025 Task 3: a Multilingual and Knowledge-Aware Nonsense Identifier} 
\title{TUM-MiKaNi at SemEval-2025 Task 3: Hallucination Detection using Knowledge Retrieval, BERT, and Support Vector Regression}
\title{TUM-MiKaNi at SemEval-2025 Task 3: Towards Multilingual and Knowledge-Aware Non-factual Hallucination Identification} 

\author{Miriam Anschütz\thanks{Equal contribution} \and Ekaterina Gikalo\footnotemark[1] \and Niklas Herbster\footnotemark[1] \and {\bf 
 Georg Groh} \\
 School of Computation, Information and Technology\\
        Technical University of Munich \\ \texttt{\{\href{mailto:miriam.anschuetz@tum.de}{miriam.anschuetz}, \href{mailto:ekaterina.gikalo@tum.de}{ekaterina.gikalo}, \href{mailto:niklas.herbster@tum.de}{niklas.herbster}\}@tum.de}}

\begin{document}
\maketitle
\begin{abstract}
Hallucinations are one of the major problems of LLMs, hindering their trustworthiness and deployment to wider use cases. However, most of the research on hallucinations focuses on English data, neglecting the multilingual nature of LLMs. This paper describes our submission to the  "\textit{SemEval-2025 Task-3 — Mu-SHROOM, the Multilingual Shared-task on Hallucinations and Related Observable Overgeneration Mistakes}". We propose a two-part pipeline that combines retrieval-based fact verification against Wikipedia with a BERT-based system fine-tuned to identify common hallucination patterns. Our system achieves competitive results across all languages, reaching top-10 results in eight languages, including English. Moreover, it supports multiple languages beyond the fourteen covered by the shared task. This multilingual hallucination identifier can help to improve LLM outputs and their usefulness in the future.

\end{abstract}

\section{Introduction}
Hallucinations are unwanted parts in the LLM outputs that are either not aligned with the source document (intrinsic) or non-factual in terms of world knowledge (extrinsic) \citep{narayanan-venkit-hallucination-overview}. These over-generations are a severe problem in NLP research, and their detection and mitigation are studied widely \citep{rashad-etal-2024-factalign, islam2025-multilingual-hallucinations}.
However, most hallucination research focuses on English data. Therefore, the "\textit{SemEval-2025 Task-3 — Mu-SHROOM, the Multilingual Shared-task on Hallucinations and Related Observable Overgeneration Mistakes}" provides hallucination annotations in fourteen languages and asks the participants to create multilingual hallucination detectors \citep{vazquez-etal-2025-mu-shroom}. Their data comes from open-source instruction-tuned LLMs, generated in a QA setting, and humans annotated each token to determine whether it belongs to a hallucinated phrase. Each token receives two different labels: A binary hard label, obtained as a majority vote between the annotators, and a soft label $\in [0,1]$ given by the proportion of annotators who labeled the token as a hallucination.

\begin{figure}[t]
  \includegraphics[width=\columnwidth]{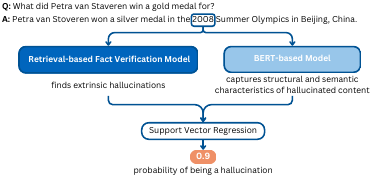}
  \caption{Given a question-answer pair, each token in the answer is evaluated for extrinsic hallucinations using a retrieval-based fact verification model, which compares the token against external knowledge. Simultaneously, the Bert-based Model captures structural and semantic characteristics of hallucinated content. The results from both models are then integrated using Support Vector Regression to estimate the final probability of each token being a hallucination.}
  \label{fig:vis_abstract}
\end{figure}

In this paper, we present our contribution to the shared task: a \textbf{MultilIngual and Knowledge-Aware Non-factual hallucination Identifier (MiKaNi)}. Our system is model-agnostic and supports multiple languages beyond the fourteen covered in the shared task. It annotates hallucinations on a token level, providing fine-grained information about the correctness of atomic facts. To obtain these annotations, we propose a two-part pipeline (visible in \autoref{fig:vis_abstract}): The first part is an atomic fact-checker based on the retrieval of information from Wikipedia. The second part is a fine-tuned BERT model incorporating linguistic features. The predictions from the two parts are then combined using a Support-Vector Regression model.
Our combined system achieves competitive scores across all languages (e.g., 8/44 in English, 2/33 in French).


Our code and model weights are publicly available for further usage and development\footnote{\href{https://github.com/MiriUll/MiKaNi-Towards-Multilingual-and-Knowledge-Aware-Non-factual-Hallucination-Identification}{Code on Github}, You can also test the fact checker \href{https://tools.ellngr.com/en/verify-claim}{here}!}.
\section{Background and related work}

Many different approaches exist to detect hallucinations in generated texts \citep{narayanan-venkit-hallucination-overview}. Some of them take information about the generating model into account, e.g., by analyzing attention weights or the model's logits \citep{sriramanan-eigenvalue-hallucinations}. While the logits of the last model layer were provided in this shared task, we created a model-agnostic detection mechanism that does not require any information about the underlying LLM.

Intrinsic hallucinations occur when the model outputs deviate from the source document. This is often the case in summarization or RAG-based tasks that provide long contexts for the model generations \citep{ravi2024-lynx-detector}. In contrast, in the Mu-SHROOM data, the model only receives a question that it has to answer based on its internal knowledge. Therefore, the main interest point of our hallucination identifier is to find extrinsic hallucinations. This entails a fact-checking or verification objective. The most popular way to check the facts in generated texts is to compare them against world knowledge covered in Wikipedia or knowledge graphs \citep{min-etal-2023-factscore}. Thus, the first part of our system builds upon this, comparing the atomic facts against retrieved data from the English Wikipedia. Previous works solved this comparison by predicting the entailment \citep{rawte-factoid} or the edit operations that are necessary to transfer the retrieved information into the model generations \citep{mishra2024-FAVA}. 



\section{System overview}
Our approach integrates the strengths of two complementary submodels, whose outputs are combined in a third model to generate the final hallucination prediction. The first submodel is designed to detect extrinsic hallucinations by retrieving relevant Wikipedia facts and using them to assess token-level hallucination probabilities. The second submodel is BERT-based and trained on token-level hallucination probability-annotated data. It focuses on identifying 
common hallucination patterns.
The outputs of both submodels, along with additional extracted features, are fed into a Support Vector Regression (SVR) model, producing the final combined hallucination score. Details about this architecture are presented in \autoref{fig:vis_abstract}.

\subsection{Retrieval-based Fact Verification Model (RFVM)}
The RFVM is a multi-step pipeline detecting hallucinations in LLM outputs by leveraging Wikipedia as a factual reference source. Given a question-answer (QA) pair, the model extracts atomic facts from the answer, retrieves and ranks relevant Wikipedia passages, and ultimately uses the retrieved evidence to assess the factuality of each token in the answer.
The model's architecture and the GPT-4o system prompts are presented in the Appendix in \autoref{fig:wikipedia_retrieval} and \autoref{app:prompts}, respectively.

\subsubsection{Pipeline Description}
\paragraph{Atomic Fact Extraction}
The first step involves breaking down the LLM-generated answer into atomic facts. An atomic fact is a self-contained statement that can be independently verified as true or false \citep{min-etal-2023-factscore}. We use GPT-4o to extract these facts in a few-shot setting.

Since the subsequent retrieval and ranking processes rely on English text, the atomic facts are translated into English during the fact extraction process. 

\paragraph{Search Term Extraction}
Once atomic facts are obtained, we extract search terms that will be used to retrieve relevant Wikipedia articles. This is done via LLM-based prompting, where GPT-4o generates a set of search terms most relevant to each atomic fact. These search terms serve as the query inputs for Wikipedia-based fact retrieval.

\paragraph{Wikipedia Fact Retrieval}
This is built upon the "Retrieval-Augmented Evaluation Pipeline" by \citet{ellinger2024evaluating} and consists of three core steps: (1) search, (2) rank, and (3) select. 

\paragraph{(1) Search} The search phase is an iterative process in which each atomic fact and its associated search terms are processed. Each search term is queried using the Wikipedia API. If a Wikipedia page with an exact title match exists, the process continues with the next steps directly. If no exact match is found, Wikipedia’s built-in suggestion mechanism is used to retrieve up to two alternative pages that may be relevant to the search term. To improve efficiency, 
searches are cached.

\paragraph{(2) Sentence Ranking}
Once a Wikipedia page has been retrieved, its sentences are processed and ranked based on their relevance to the corresponding atomic fact. We first apply co-reference resolution to the entire page to ensure consistency at the sentence level. The text is then split into individual sentences, creating a structured content representation. Finally, each sentence is ranked using the BM25 retrieval algorithm.

\paragraph{(3) Evidence Selection}
The final step in the retrieval process selects the most relevant sentences based on one of two strategies:
\begin{itemize}
    \item \textbf{Top-$n$ Selection}: The top-$n$ most relevant sentences, as determined by BM25 ranking, are selected.
    \item \textbf{Maximal Marginal Relevance (MMR) Selection}: Selects highly relevant sentences while ensuring diversity.
\end{itemize}

\paragraph{Fact Verification and Hallucination Prediction}
The output of the fact retrieval module is a structured list of dictionaries, where each entry consists of an atomic fact and its most relevant Wikipedia evidence.

This structured evidence, along with the original QA pair, is then passed to GPT-4o for hallucination detection. The process begins by splitting the generated answer into individual sentences. Each sentence is evaluated separately, with the full list of retrieved Wikipedia facts and the original question provided as context. GPT-4o estimates the probability of each token being hallucinated based on this evidence. To accelerate inference, sentence-level concurrency is employed, allowing multiple sentences to be processed in parallel.

\paragraph{Final Aggregation}
Once all individual hallucination predictions are obtained from the LLM, the results are aggregated into a single final hallucination probability distribution over the entire answer.

\subsection{BERT-based Model (BM)}
\begin{figure}[t]
  \includegraphics[width=\columnwidth]{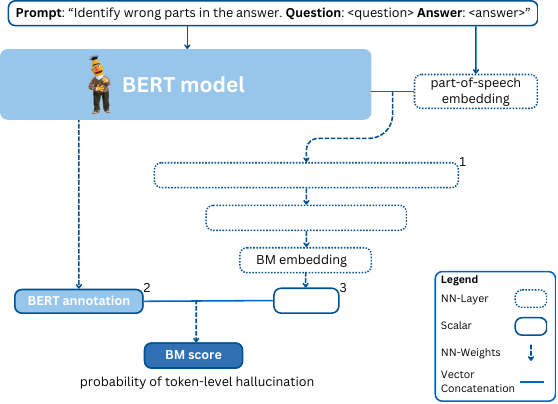}
  \caption{BERT-based Model architecture: BERT is prompted with the instruction, question, and answer. It is then enhanced with the answer's part-of-speech embedding and processed through several fully connected layers to obtain the final token hallucination score.}
  \label{fig:bm_arch}
\end{figure}

The BERT-based Model (BM) builds on a \href{https://huggingface.co/google-bert/bert-base-multilingual-cased}{pre-trained BERT model} \citep{devlin-bert} to detect hallucinations in the generated text. The model processes a structured prompt containing an instruction, a question, and an answer, as illustrated in Figure \ref{fig:bm_arch}.

First, the output embedding from BERT is extracted and concatenated with a part-of-speech (POS) embedding. 
The POS embedding is generated using SpaCy \citep{honnibal-spacy}, where each token in the answer is represented by a numerical POS tag. 
The combined BERT-POS representation is then passed through several linear layers (1), each with a ReLU activation function. The output of these layers produces an intermediate representation: the BM embedding. Additionally, the original BERT output is re-introduced into the model by processing it through a fully-connected layer to obtain a BERT annotation (2). This annotation is then concatenated with the processed BM embedding (3). The resulting representation is subsequently processed through a final fully connected layer, which computes the BM probability of a token being a hallucination (BM score).

\subsection{Support Vector Regression model (SVRM)}
\begin{figure}[t]
  \includegraphics[width=\columnwidth]{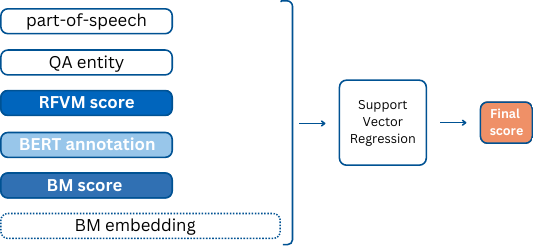}
  \caption{SVR Model architecture: POS embedding, QA entity, RFVM score, BERT annotation, BM score, and BM embedding are concatenated and processed as input for the Support Vector Regression.}
  \label{fig:svrm_arch}
\end{figure}

The final hallucination score is determined using a Support Vector Regression model \citep{drucker-SVR} that combines various linguistic and contextual features. 
The ensemble of different models and features, like neural embeddings, fine-tuned models, or linguistic features, has shown good results in shared task submissions before \citep{liu-shroom-ensemble, anschutz-GermEval-SVR-combination}. Thus, we opted for a similar combined approach.
Our features include POS tags, question-answer entity matches, and outputs from previous models: the RFVM score, the BERT annotation, the BM score, and the BM embedding, as depicted in \autoref{fig:svrm_arch}.

The question-answer entity feature is a binary value assigned to each token in the answer, indicating if it is part of a named entity in the question. Named entity annotations are obtained using SpaCy or Stanza \citep{qi2020stanza}, and the identified entities are matched to BERT tokens.
This feature helps to recall non-hallucinated tokens, as we found that many true hallucinations are named entities (see \autoref{fig:pos_vs_hall} in the appendix).

All features are concatenated into a unified representation, which the SVR model uses to learn a regression function that predicts the final hallucination probability for each token.

\subsection{Experimental setup}
\subsubsection{RFVM Setup}
The RFVM model leverages GPT-4o as LLM for various prompting-based tasks. To maximize the model's performance, each step in the pipeline is guided by a carefully crafted system prompt along with a set of few-shot examples. The prompts are presented in \autoref{app:prompts}. Prompting is employed in three key stages: atomic fact extraction, search term generation, and final hallucination prediction. 

\paragraph{System Prompt Structure}
The system prompts across all three stages follow a consistent structure. Each prompt begins with an introductory paragraph that provides contextual background on the task. This is followed by a section containing detailed task instructions, specifying the expected model behavior in a precise and structured manner. Finally, the prompt explicitly defines the expected input and output format, ensuring that the model generates responses in a structured and processable form.

\paragraph{Prompting for Atomic Fact Extraction}
For atomic fact extraction, the model is provided with a system prompt that includes a structured task description along with three illustrative examples. Two of these examples are in English, while one is in Spanish, preparing the model for multilingual inputs. These few-shot examples serve to clearly define the task of breaking down complex answers into simple, verifiable statements.

\paragraph{Prompting for Search Term Generation}
In the search term generation step, the system prompt includes two examples in English. During the atomic fact extraction, all facts are translated into English. Thus, at this stage, all input data is strictly in English to ensure consistency throughout the retrieval process.  Since the quality of search terms directly impacts retrieval success, the prompt was iteratively improved using trial and error. Extracting search terms that lead to a valid Wikipedia page hit is a non-trivial task requiring precise guidance. The system prompt for this step contains 13 detailed instructional steps to ensure the generated search terms maximize retrieval accuracy.

\paragraph{Prompting for Hallucination Prediction}
For the final hallucination prediction step, a single English QA pair from the Mu-SHROOM validation set is used as an example. Including only one example is primarily driven by computational and cost-efficiency considerations. Despite this limitation, the example is carefully chosen to demonstrate the hallucination detection task.

\paragraph{Evidence Selection}
For evidence selection, we opted for MMR to ensure the diversity of search results while also maintaining  the relevance of facts. As MMR parameters we use top$_n = 4$ and $\lambda = 0.7$ to balance relevance and diversity.

\subsubsection{BM training}
The BM was trained on a multilingual dataset that consists of Mu-SHROOM SemEval validation datasets for English, German, French, Finnish, Swedish, Italian, and Spanish \citep{vazquez-etal-2025-mu-shroom}. The dataset is divided into training, validation, and test sets, following an 80/10/10 split.

The training was conducted over 10 epochs, using separate learning rates for fine-tuning BERT ($5\mathrm{e}{-5}$) and training the fully connected (FC) layers ($3\mathrm{e}{-4}$). A batch size of 1 was used to account for the token-based processing. The learning rate for the FC layers was multiplied by a factor of 0.5 if no improvement was observed for three consecutive epochs. To preserve initial features, the first three layers of BERT were frozen during training. 

We use Mean Squared Error (MSE) as our loss, with triple weighting applied to labels where the hallucination probability was greater than zero. The loss was calculated and backpropagated for each token in the answer while the computational graph was retained. After processing all tokens, the prediction variance was calculated, scaled by a regularization rate of 0.9, and then backpropagated to penalize uniform outputs. After the initial training, the model was fine-tuned for three additional epochs with a lower regularization rate of 0.5 while keeping other hyperparameters unchanged.

\subsubsection{SVR training}
The regression model was trained on the soft labels, using all languages in the Mu-SHROOM training data. The dataset was divided into training and validation sets using a 90/10 split.

The SVR model was trained with a regularization parameter \(C = 10\) to increase sensitivity to errors. Non-hallucination samples (with soft labels of 0) were weighted at 0.01, while all other samples were given a weight of 100. For POS tagging, SpaCy was used for all languages except Arabic, Hindi, Czech, Basque, and Persian, where Stanza was applied. Additionally, word spans were merged if the distance between words was less than three and the probability difference did not exceed 15\%, with the higher hallucination probability being retained for the combined span.




\section{Results}
The Hallucinations were evaluated using intersection-over-union (IoU) for hard labels and Spearman correlation (Cor) for soft labels. IoU measures the overlap between hallucination spans, while Cor assesses the correlation between predicted and reference probabilities \citep{vazquez-etal-2025-mu-shroom}.

The submission results, including IoU and Cor scores along with the corresponding ranks, are presented in \autoref{tab:results_ranks}. The shared task organizers provided three baselines: a \textit{neural baseline}, a \textit{mark all} baseline that labels everything as hallucinations, and a \textit{mark none} baseline. Our system outperformed these baselines in all languages except Chinese, where the mark-all baseline performed slightly better. This shows that our combined approach is successful.
However, the performances vary across languages. This could be due to the underlying data, as the best-performing systems per language also show a great range of IoU scores (between 0.53 in Spanish and 0.79 in Italian). Another factor is the dependence on SpaCy and Stanza annotations, as their quality may decrease for certain languages. Nevertheless, our system shows a good generalization to unseen test-only languages like Catalan and Farsi.

Tables \ref{tab:individual_results_base_iou} and \ref{tab:individual_results_base_cor} provide test-set scores for our two subsystems separately. The BERT-based model tends to outperform the RFVM in most languages. However, the BM still benefits from further annotations in the RFVM results, resulting in higher scores for the ensembled models. A further analysis of language-specific behavior and a more qualitative analysis is provided in \autoref{app:futher_analysis}.


\begin{table}[]
    \centering
    \begin{tabular}{lccr} \toprule
        \textbf{Language} & \textbf{IoU} $\uparrow$ & \textbf{Cor} & \textbf{Rank} \\ 
        \midrule
        Italian & 0.6787 & 0.5388 & 12/31 \\
        \textbf{French} & 0.6314 & 0.5157 & 2/33 \\
        \textbf{Finnish} & 0.6267 & 0.5751 & 5/30 \\
        \textit{\textbf{Catalan}} & 0.5971 & 0.5551 & 7/24 \\
        \textbf{Swedish} & 0.5886 & 0.3930 & 6/30 \\
        Hindi & 0.5835 & 0.4964 & 12/27 \\
        German & 0.5569 & 0.5088 & 10/31 \\
        \textit{Farsi} & 0.5465 & 0.4238 & 11/26 \\
        \textbf{English} & 0.5249 & 0.5363 & 8/44 \\
        \textit{\textbf{Basque}} & 0.5237 & 0.4709 & 7/26 \\
        Arabic & 0.4778 & 0.5114 & 14/32 \\
        \textbf{Chinese} & 0.4735 & 0.4095 & 9/29 \\
        \textit{Czech} & 0.3874 & 0.3738 & 12/26 \\
        Spanish & 0.3739 & 0.5027 & 14/35 \\
        \bottomrule
    \end{tabular}
    \caption{Results across languages, sorted by IoU scores. All languages, except Chinese, outperformed the baselines. Languages where our submission is in the top 10 of all submitted systems are bolded. Languages in \textit{italic} are test-only languages without available training data.}
    \label{tab:results_ranks}
\end{table}

\section{Conclusion}
In this paper, we present MiKaNi, a multilingual and knowledge-aware hallucination identification system that achieved competitive performance in the Mu-SHROOM shared task. Our system combines fact verification against external sources with a BERT-based system fine-tuned to detect common hallucination patterns. The system uses the same architecture for all languages and LLM outputs, making it strongly multilingual and model-agnostic, generalizing well on the unseen test set languages. The language capacities for further languages are only limited by the availability of SpaCy or Stanza annotations and their support in multilingual BERT. In future work, we will try to make the model even more flexible by testing open-source fact verification models instead of relying on OpenAI's GPT-4o.

\section{Limitations}
Our Retrieval-based Fact Verification Model (RFVM) heavily relies on prompting GPT-4o to obtain the atomic facts and search terms and to perform the overall hallucination prediction. 
While this API is easy to use and GPT-4o generates high-quality responses, relying on closed-source models limits the reusability of our approach for other researchers, particularly those with limited financial resources. For future work, we plan to experiment with open-source and more lightweight models to reduce this barrier. 

Another limitation of our pipeline is the high latency during inference due to the modular and sequential design. A QA pair has to be processed by our RFVM, including a retrieval process against the Wikipedia API and multiple calls to the OpenAI API. While the RFVM and the BM predictions can run in parallel, the SVRM depends on both outputs and, thus, has to wait for these results. During development, we focussed our efforts on a good performance in as many languages as possible. However, for deployment of our model in an LLM interface, the individual pipeline steps would have to be improved for efficiency.

\section*{Acknowledgments}
The Retrieval-based Fact Verification Model (RFVM) was inspired by a Master's thesis by \citet{ellinger2024evaluating}. We thank Lukas for sharing his code with us and for his continued support.

Some parts of this paper were written with the help of AI assistant tools in the form of ChatGPT. All AI-generated contents were thoroughly revised by the authors.


\begin{thebibliography}{16}
\providecommand{\natexlab}[1]{#1}

\bibitem[{Ansch{\"u}tz and Groh(2022)}]{anschutz-GermEval-SVR-combination}
Miriam Ansch{\"u}tz and Georg Groh. 2022.
\newblock \href {https://aclanthology.org/2022.germeval-1.4/} {{TUM} social
  computing at {G}erm{E}val 2022: Towards the significance of text statistics
  and neural embeddings in text complexity prediction}.
\newblock In \emph{Proceedings of the GermEval 2022 Workshop on Text Complexity
  Assessment of German Text}, pages 21--26, Potsdam, Germany. Association for
  Computational Linguistics.

\bibitem[{Devlin et~al.(2019)Devlin, Chang, Lee, and Toutanova}]{devlin-bert}
Jacob Devlin, Ming-Wei Chang, Kenton Lee, and Kristina Toutanova. 2019.
\newblock \href {https://doi.org/10.18653/v1/N19-1423} {{BERT}: Pre-training of
  deep bidirectional transformers for language understanding}.
\newblock In \emph{Proceedings of the 2019 Conference of the North {A}merican
  Chapter of the Association for Computational Linguistics: Human Language
  Technologies, Volume 1 (Long and Short Papers)}, pages 4171--4186,
  Minneapolis, Minnesota. Association for Computational Linguistics.

\bibitem[{Drucker et~al.(1996)Drucker, Burges, Kaufman, Smola, and
  Vapnik}]{drucker-SVR}
Harris Drucker, Chris J.~C. Burges, Linda Kaufman, Alex Smola, and Vladimir
  Vapnik. 1996.
\newblock Support vector regression machines.
\newblock In \emph{Proceedings of the 10th International Conference on Neural
  Information Processing Systems}, NIPS'96, page 155–161, Cambridge, MA, USA.
  MIT Press.

\bibitem[{Honnibal et~al.(2020)Honnibal, Montani, Landeghem, and
  Boyd}]{honnibal-spacy}
Matthew Honnibal, Ines Montani, Sofie~Van Landeghem, and Adriane Boyd. 2020.
\newblock \href {https://doi.org/10.5281/zenodo.1212303} {spacy:
  Industrial-strength natural language processing in python}.

\bibitem[{Liu et~al.(2024)Liu, Shi, Zhang, and Huang}]{liu-shroom-ensemble}
Wei Liu, Wanyao Shi, Zijian Zhang, and Hui Huang. 2024.
\newblock \href {https://doi.org/10.18653/v1/2024.semeval-1.253}
  {{HIT}-{MI}{\&}{T} lab at {S}em{E}val-2024 task 6: {D}e{BERT}a-based
  entailment model is a reliable hallucination detector}.
\newblock In \emph{Proceedings of the 18th International Workshop on Semantic
  Evaluation (SemEval-2024)}, pages 1788--1797, Mexico City, Mexico.
  Association for Computational Linguistics.

\bibitem[{{Lukas Ellinger}(2024)}]{ellinger2024evaluating}
{Lukas Ellinger}. 2024.
\newblock \href {https://github.com/lukasellinger/evaluating-word-definitions}
  {Retrieval-augmented evaluation: Assessing the factuality of word definitions
  using wikipedia}.
\newblock Master's thesis, {Technical University of Munich}.
\newblock Advised and supervised by Miriam Ansch{\"u}tz and Georg Groh.

\bibitem[{Min et~al.(2023)Min, Krishna, Lyu, Lewis, Yih, Koh, Iyyer,
  Zettlemoyer, and Hajishirzi}]{min-etal-2023-factscore}
Sewon Min, Kalpesh Krishna, Xinxi Lyu, Mike Lewis, Wen-tau Yih, Pang Koh, Mohit
  Iyyer, Luke Zettlemoyer, and Hannaneh Hajishirzi. 2023.
\newblock \href {https://doi.org/10.18653/v1/2023.emnlp-main.741}
  {{FA}ct{S}core: Fine-grained atomic evaluation of factual precision in long
  form text generation}.
\newblock In \emph{Proceedings of the 2023 Conference on Empirical Methods in
  Natural Language Processing}, pages 12076--12100, Singapore. Association for
  Computational Linguistics.

\bibitem[{Mishra et~al.(2024)Mishra, Asai, Balachandran, Wang, Neubig,
  Tsvetkov, and Hajishirzi}]{mishra2024-FAVA}
Abhika Mishra, Akari Asai, Vidhisha Balachandran, Yizhong Wang, Graham Neubig,
  Yulia Tsvetkov, and Hannaneh Hajishirzi. 2024.
\newblock \href {https://openreview.net/forum?id=dJMTn3QOWO} {Fine-grained
  hallucination detection and editing for language models}.
\newblock In \emph{First Conference on Language Modeling}.

\bibitem[{Narayanan~Venkit et~al.(2024)Narayanan~Venkit, Chakravorti, Gupta,
  Biggs, Srinath, Goswami, Rajtmajer, and
  Wilson}]{narayanan-venkit-hallucination-overview}
Pranav Narayanan~Venkit, Tatiana Chakravorti, Vipul Gupta, Heidi Biggs, Mukund
  Srinath, Koustava Goswami, Sarah Rajtmajer, and Shomir Wilson. 2024.
\newblock \href {https://doi.org/10.18653/v1/2024.emnlp-main.375} {An audit on
  the perspectives and challenges of hallucinations in {NLP}}.
\newblock In \emph{Proceedings of the 2024 Conference on Empirical Methods in
  Natural Language Processing}, pages 6528--6548, Miami, Florida, USA.
  Association for Computational Linguistics.

\bibitem[{Qi et~al.(2020)Qi, Zhang, Zhang, Bolton, and Manning}]{qi2020stanza}
Peng Qi, Yuhao Zhang, Yuhui Zhang, Jason Bolton, and Christopher~D. Manning.
  2020.
\newblock \href {https://nlp.stanford.edu/pubs/qi2020stanza.pdf} {Stanza: A
  {Python} natural language processing toolkit for many human languages}.
\newblock In \emph{Proceedings of the 58th Annual Meeting of the Association
  for Computational Linguistics: System Demonstrations}.

\bibitem[{Rashad et~al.(2024)Rashad, Zahran, Amin, Abdelaal, and
  Altantawy}]{rashad-etal-2024-factalign}
Mohamed Rashad, Ahmed Zahran, Abanoub Amin, Amr Abdelaal, and Mohamed
  Altantawy. 2024.
\newblock \href {https://doi.org/10.18653/v1/2024.trustnlp-1.8} {{F}act{A}lign:
  Fact-level hallucination detection and classification through knowledge graph
  alignment}.
\newblock In \emph{Proceedings of the 4th Workshop on Trustworthy Natural
  Language Processing (TrustNLP 2024)}, pages 79--84, Mexico City, Mexico.
  Association for Computational Linguistics.

\bibitem[{Ravi et~al.(2024)Ravi, Mielczarek, Kannappan, Kiela, and
  Qian}]{ravi2024-lynx-detector}
Selvan~Sunitha Ravi, Bartosz Mielczarek, Anand Kannappan, Douwe Kiela, and
  Rebecca Qian. 2024.
\newblock \href {https://arxiv.org/abs/2407.08488} {Lynx: An open source
  hallucination evaluation model}.
\newblock \emph{Preprint}, arXiv:2407.08488.

\bibitem[{Rawte et~al.(2024)Rawte, Tonmoy, Rajbangshi, Nag, Chadha, Sheth, and
  Das}]{rawte-factoid}
Vipula Rawte, S.~M. Towhidul~Islam Tonmoy, Krishnav Rajbangshi, Shravani Nag,
  Aman Chadha, Amit~P. Sheth, and Amitava Das. 2024.
\newblock \href {https://doi.org/10.48550/arXiv.2403.19113} {Factoid: Factual
  entailment for hallucination detection}.
\newblock \emph{CoRR}, abs/2403.19113.

\bibitem[{Sriramanan et~al.(2024)Sriramanan, Bharti, Sadasivan, Saha,
  Kattakinda, and Feizi}]{sriramanan-eigenvalue-hallucinations}
Gaurang Sriramanan, Siddhant Bharti, Vinu~Sankar Sadasivan, Shoumik Saha,
  Priyatham Kattakinda, and Soheil Feizi. 2024.
\newblock \href
  {https://proceedings.neurips.cc/paper_files/paper/2024/file/3c1e1fdf305195cd620c118aaa9717ad-Paper-Conference.pdf}
  {Llm-check: Investigating detection of hallucinations in large language
  models}.
\newblock In \emph{Advances in Neural Information Processing Systems},
  volume~37, pages 34188--34216. Curran Associates, Inc.

\bibitem[{ul~Islam et~al.(2025)ul~Islam, Lauscher, and
  Glavaš}]{islam2025-multilingual-hallucinations}
Saad~Obaid ul~Islam, Anne Lauscher, and Goran Glavaš. 2025.
\newblock \href {https://arxiv.org/abs/2502.12769} {How much do llms
  hallucinate across languages? on multilingual estimation of llm hallucination
  in the wild}.
\newblock \emph{Preprint}, arXiv:2502.12769.

\bibitem[{V\'azquez et~al.(2025)V\'azquez, Mickus, Zosa, Vahtola, Tiedemann,
  Sinha, Segonne, S\'anchez-Vega, Raganato, Libovický, Karlgren, Ji, Helcl,
  Guillou, de~Gibert, Bengoetxea, Attieh, and
  Apidianaki}]{vazquez-etal-2025-mu-shroom}
Ra\'ul V\'azquez, Timothee Mickus, Elaine Zosa, Teemu Vahtola, J\"org
  Tiedemann, Aman Sinha, Vincent Segonne, Fernando S\'anchez-Vega, Alessandro
  Raganato, Jindřich Libovický, Jussi Karlgren, Shaoxiong Ji, Jindřich
  Helcl, Liane Guillou, Ona de~Gibert, Jaione Bengoetxea, Joseph Attieh, and
  Marianna Apidianaki. 2025.
\newblock \href {https://helsinki-nlp.github.io/shroom/} {Sem{E}val-2025 {T}ask
  3: {Mu-SHROOM}, the multilingual shared-task on hallucinations and related
  observable overgeneration mistakes}.

\end{thebibliography}

\appendix

\section{Further analysis and discussion}\label{app:futher_analysis}
\begin{figure}[]
  \includegraphics[width=\columnwidth]{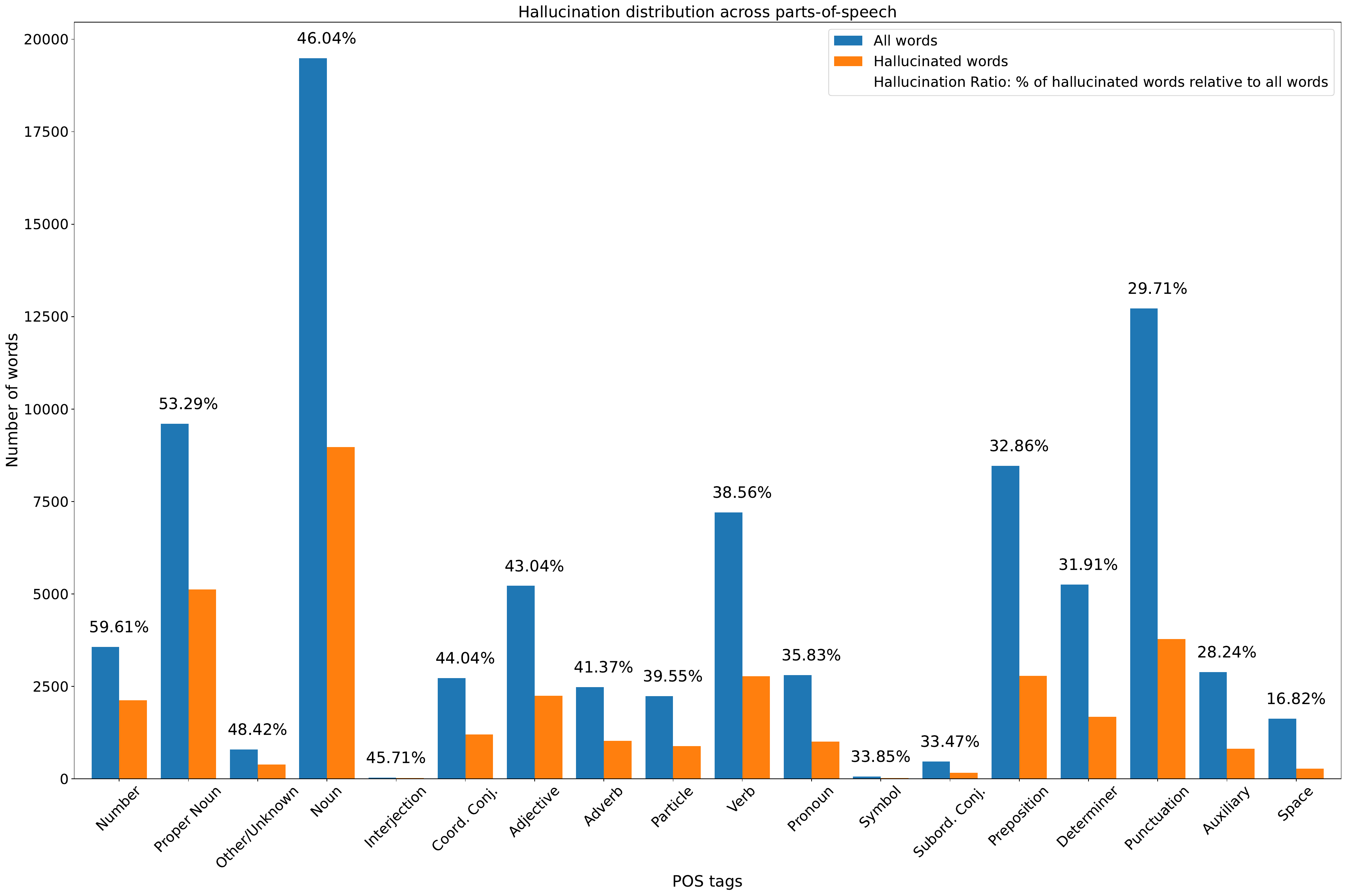}
  \caption{Comparison between part-of-speech tags and hard labels, sorted by hallucination ratio.}
  \label{fig:pos_vs_hall}
\end{figure}

\begin{figure}[]
\includegraphics[width=\columnwidth]{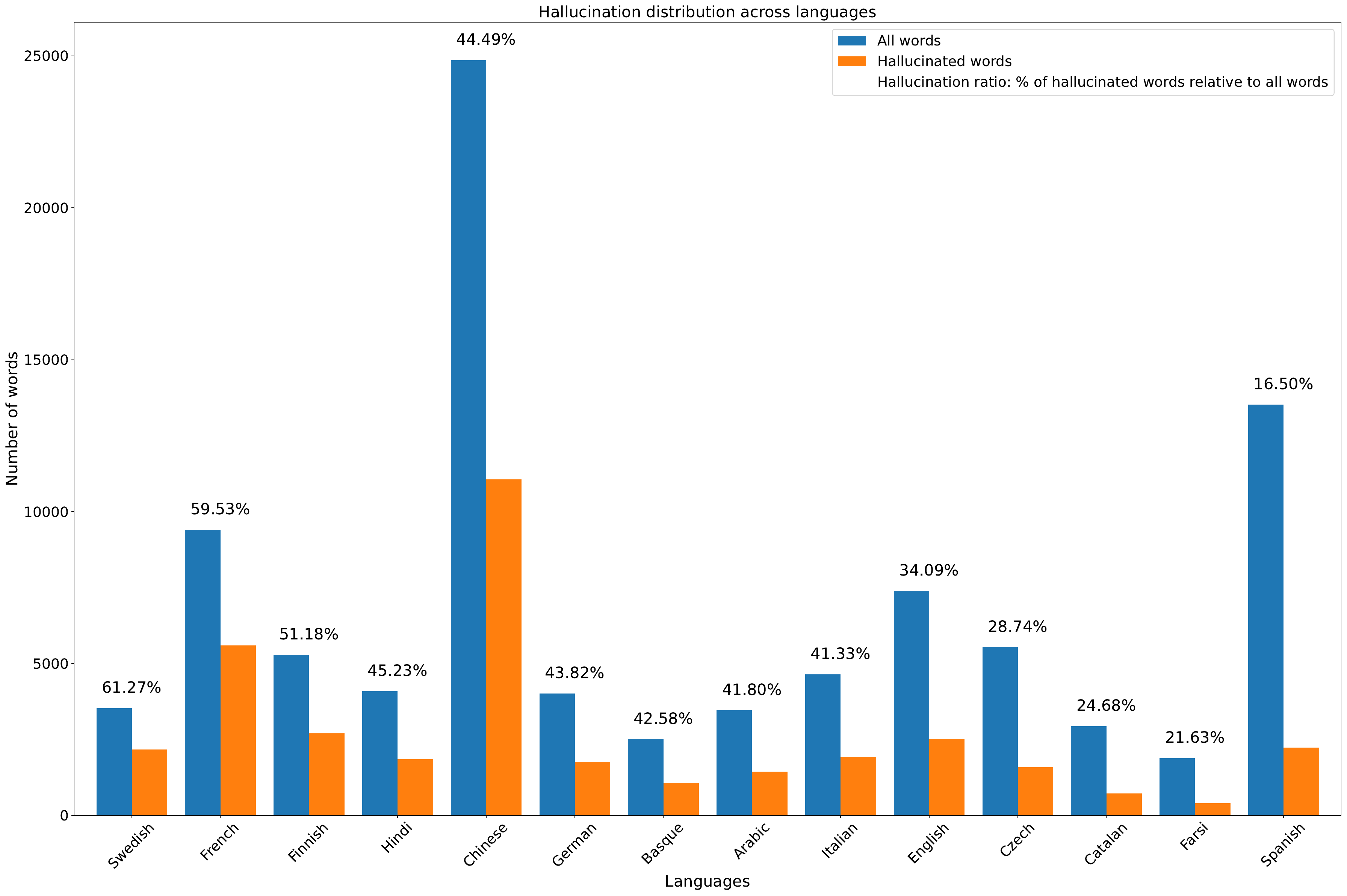}
  \caption{Comparison between language and hard labels, sorted by hallucination ratio.}
  \label{fig:lang_vs_hall}
\end{figure}

\begin{table*}
    \centering
    \begin{tabular}{lccc|ccc} \toprule\multirow{2}{*}{\textbf{Language}$\downarrow$}\ & \multicolumn{3}{c}{\textbf{Baselines}}& \multicolumn{3}{c}{\textbf{Our models}} \\
         & \textbf{mark\_none} & \textbf{mark\_all} & \textbf{neural} & \textbf{RFVM} & \textbf{BM} & \textbf{SVRM} \\
        \midrule
        Arabic & 0.0467 & 0.3613 & 0.0418 & 0.3629 & 0.4611 & \textbf{0.4778} \\ 
        \textit{Basque} & 0.0101 & 0.3671 & 0.0208 & 0.4343 & 0.4290 & \textbf{0.5237} \\
        \textit{Catalan} & 0.08 & 0.2423 & 0.0524 &  0.4411 & 0.5376 & \textbf{0.5971} \\
        \textit{Czech} & 0.13 & 0.2631 & 0.0957 & \textbf{0.3874} & 0.3816 & 0.3853 \\
        English & 0.0324 & 0.3489 & 0.0310 & 0.4650 & 0.4912 &  \textbf{0.5249} \\
        \textit{Farsi} & 0 & 0.2028 & 0.0001 &  0.3176 & 0.5315 & \textbf{0.5465} \\
        Finnish & 0 & 0.4857 & 0.0042 & 0.4868 & 0.5907 & \textbf{0.6267} \\
        French & 0 & 0.4543 & 0.0022 & 0.3435 & 0.5622 & \textbf{0.6314} \\
        German & 0.0267 & 0.3450 & 0.0318 & 0.4907 & 0.4735 & \textbf{0.5569} \\
        Hindi & 0 & 0.2711 & 0.0029 & 0.3584 & 0.5692 & \textbf{0.5835} \\
        Italian & 0 & 0.2826 & 0.0104 & 0.4618 & \textbf{0.6787} & 0.6781 \\
        Spanish & 0.0855 & 0.1853 & 0.0724 & 0.3672 & 0.3627 & \textbf{0.3739} \\
        Swedish & 0.0204 & 0.5372 & 0.0308 & 0.5298 & 0.5434 & \textbf{0.5886} \\
        Chinese & 0.02 & \textbf{0.4772} & 0.0238 & 0.2530 & 0.4490 & 0.4735 \\
        \bottomrule
    \end{tabular}
    \caption{\textbf{IoU} scores of all baselines and our RFVM, BM, and SVRM across all languages. The languages are sorted alphabetically. Test-only languages are shown in \textit{italic}, and the best submissions are bolded. All languages outperform the baselines, except the Chinese mark all baseline.}
    \label{tab:individual_results_base_iou}
\end{table*}

\begin{table*}
    \centering
    \begin{tabular}{lccc|ccc} \toprule\multirow{2}{*}{\textbf{Language}$\downarrow$}\ & \multicolumn{3}{c}{\textbf{Baselines}}& \multicolumn{3}{c}{\textbf{Our models}} \\
         & \textbf{mark\_none} & \textbf{mark\_all} & \textbf{neural} & \textbf{RFVM} & \textbf{BM} & \textbf{SVRM} \\
        \midrule
        Arabic & 0.0067 & 0.0067 & 0.1190 & 0.2369 & 0.4947 & \textbf{0.5114} \\ 
        \textit{Basque} & 0 & 0 & 0.1004 & 0.3975 & \textbf{0.4996} & 0.4709 \\
        \textit{Catalan} & 0.06 & 0.06 & 0.0645 &  0.4626 & 0.4796 & \textbf{0.5551} \\
        \textit{Czech} & 0.1 & 0.1 & 0.0533 & 0.3738 & 0.4151 & \textbf{0.4580} \\
        English & 0 & 0 & 0.1190 & 0.4567 &  \textbf{0.5472} & 0.5363 \\
        \textit{Farsi} & 0.01 & 0.01 & 0.1078 & 0.3253 & \textbf{0.4762} & 0.4238 \\
        Finnish & 0 & 0 & 0.0924 & 0.3821 & 0.5592 & \textbf{0.5751} \\
        French & 0 & 0 & 0.0208 & 0.3006 & 0.4730 & \textbf{0.5157} \\
        German & 0.0133 & 0.0133 & 0.1073 & 0.4786 & 0.4547 & \textbf{0.5088} \\
        Hindi & 0 & 0 & 0.1429 & 0.3336 & \textbf{0.5273} & 0.4964 \\
        Italian & 0 & 0 & 0.0800 & 0.4803 & 0.5388 & \textbf{0.6233} \\
        Spanish & 0.0132 & 0.0132 & 0.0359 & 0.4312 & 0.4557 & \textbf{0.5027} \\
        Swedish & 0.0136 & 0.0136 & 0.0968 & 0.3543 & 0.3889 & \textbf{0.3930} \\
        Chinese & 0 & 0 & 0.0883 & 0.1756 & \textbf{0.4676} & 0.4095 \\
        \bottomrule
    \end{tabular}
    \caption{\textbf{Cor} scores of all baselines and our RFVM, BM, and SVRM across all languages. The languages are sorted alphabetically. Test-only languages are shown in \textit{italic}, and the best submissions are bolded.}
    \label{tab:individual_results_base_cor}
\end{table*}


Figures \ref{fig:pos_vs_hall} and \ref{fig:lang_vs_hall} illustrate the distribution of hallucinated words across part-of-speech categories and languages, respectively. Notably, Swedish exhibits the highest number of hallucinated words, while Spanish has the lowest. The majority of hallucinations occur in numbers, proper nouns, and nouns.

Spanish exhibits low IoU scores, ranking among the lowest-performing languages alongside Arabic. As shown in \autoref{tab:individual_results_base_iou}, both RFVM and BM performed poorly on Spanish, resulting in similarly low performance for SVRM, which recorded the lowest IoU scores among all languages. The underperformance of RFVM and BM may be due to the high word count in Spanish samples (see \autoref{fig:lang_vs_hall}), as both models struggle with long inputs. Additionally, BM achieved a lower IoU score than RFVM on Spanish (see \autoref{tab:individual_results_base_iou}), likely due to the very low hallucination content, as BM tends to overpredict hallucinations.

Chinese underperformed relative to the baseline, as both RFVM and BM struggle with long inputs. Chinese samples contain relatively long outputs, comparable to other languages (see \autoref{fig:lang_vs_hall}). RFVM, in particular, achieved the lowest IoU score on Chinese (see \autoref{tab:individual_results_base_iou}). This may also be attributed to the challenges of translating Chinese into English. Translation can introduce ambiguities, modify sentence structures, or obscure contextual meaning, making it more difficult for RFVM to retrieve precise matches from English Wikipedia.


On the Italian dataset, BM slightly outperformed SVRM. Since the Italian data does not exhibit significant deviations, and SVRM heavily relies on BM, this result may indicate the performance ceiling of both approaches. It also suggests the need to incorporate additional metrics into SVRM to improve its predictions.

\begin{table*}[ht]
    \centering
    \begin{tabularx}{\textwidth}{lXX} \toprule
        \textbf{Language} & \textbf{Ground truth} & \textbf{SVR prediction} \\
        \midrule
        \vspace{8pt}
        Catalan & La pel·lícula Faster, Pussycat! Kill! Kill! \textcolor{red}{no té narrador}. És una \textcolor{red}{pel·lícula muda}, de manera que \textcolor{red}{no hi ha veu en off explicant la història}. & La pel·lícula Faster, Pussycat! Kill! Kill! \textcolor{red}{no té} narrador. \textcolor{red}{És una pel·lícula muda}, de manera que \textcolor{red}{no hi} \textcolor{red}{ha} \textcolor{red}{veu} \textcolor{red}{en} \textcolor{red}{off explicant la} \textcolor{red}{història}.\\
        \vspace{8pt}
        German & Die griechische Ägäis-Insel \textcolor{red}{Angista} gehört zu den \textcolor{red}{Nördlichen Sporaden}. & Die griechische Ä\textcolor{red}{gäis}-Insel \textcolor{red}{Angista} gehört zu den \textcolor{red}{Nördlichen Sporaden}. \\
        Swedish & År 2008 var det \textcolor{red}{1 357 600} invånare i Dourbies. \textcolor{red}{Detta är en ökning med 10 000 invånare sedan 2007. Befolkningen ökade med 21,5\% under de senaste 5 åren}. & År 2008 var det \textcolor{red}{1 357 600} \textcolor{red}{invånare} i Dourbies\textcolor{red}{. Detta} \textcolor{red}{är en ökning} \textcolor{red}{med 10 000} \textcolor{red}{invånare sedan} \textcolor{red}{2007. Befolkningen ökade med 21,5\% under de senaste 5} \textcolor{red}{åren}.\\
        \bottomrule
    \end{tabularx}
    \caption{Randomly selected annotation examples. Hallucinations are highlighted in red. Our SVR model sometimes annotates too many tokens, but covers the right spans in general.}
    \label{tab:examples}
\end{table*}

Overall, Tables \ref{tab:individual_results_base_iou} and \ref{tab:individual_results_base_cor} show that the ensemble SVRM model benefits from both models. For example, the German IoU scores of RFVM and BM are close to one another at $0.4907$ and $0.4735$, respectively. However, if the two models are combined in the SVRM, the score increases to $0.5569$. A similar behavior can be seen in Basque, English, and Swedish.

\begin{figure*}[ht]
    \centering
    \includegraphics[width=\textwidth]{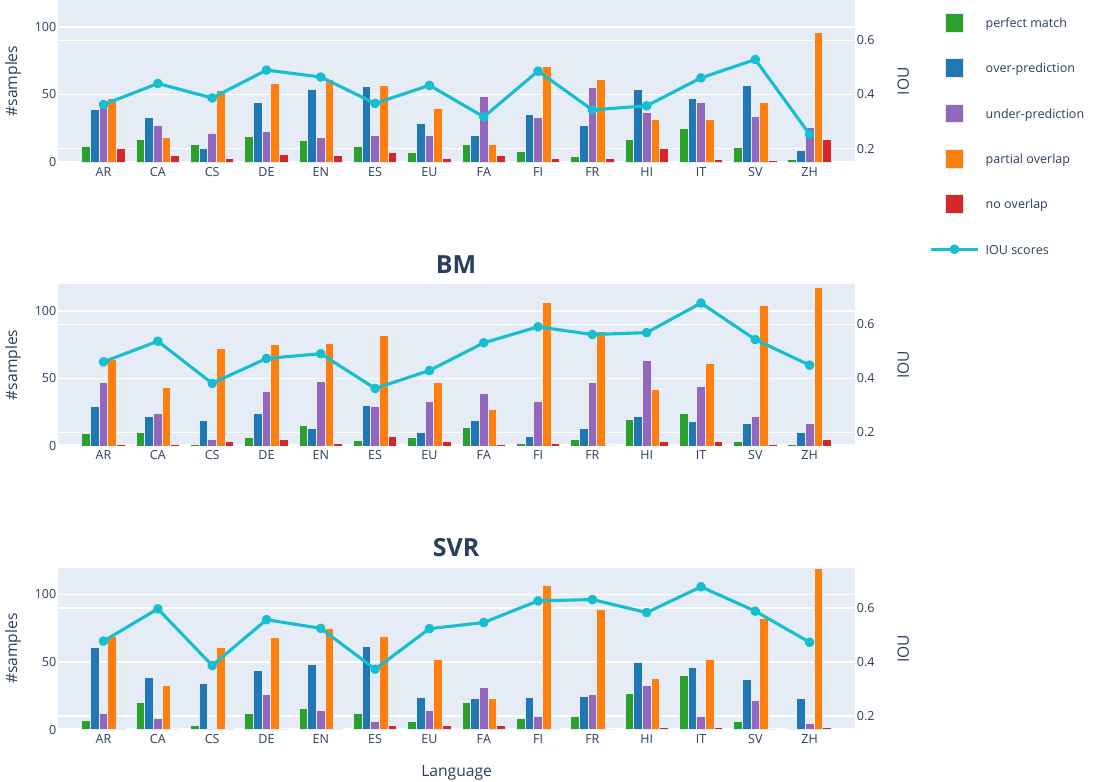}
    \caption{Caption}
    \label{fig:qual_analysis}
\end{figure*}

To further analyze the shortcomings of our models, we investigate the mispredicted spans. \autoref{fig:qual_analysis} shows the number of samples per model that have a perfect overlap of hallucination span annotations, a partial overlap, or no overlaps at all. The RFVM seems to strike a good balance between over-prediction, i.e., predicting too many false positives, and under-predictions that miss some spans. However, it is also the model with the most failures, mostly due to no annotations at all. Combining the predictions in the SVR model results in a higher rate of over-predictions, more perfect matches, and fewer failure cases. Some examples are shown in \autoref{tab:examples}.

\section{GPT-4o system prompts}\label{app:prompts}
\subsection{Atomic Fact Extraction System Prompt}
\small
\begin{flushleft}
\begin{verbatim}
You are a fact extractor. Your task is to split 
the answer to a given question into atomic facts.
Atomic facts are concise, self-contained statements
that are free from ambiguity or dependency on
context beyond the statement itself. Each fact
should be clear, stand alone, and should not assume
any implicit understanding from other facts or 
the question.

  When performing the task, adhere to the following 
  principles:

  ### Input:
  The input will be a dictionary with the following
  structure:
  {
    "question": "The question providing context.",
    "answer": "The complex answer to be broken 
               down into atomic facts."
  }

  ### Output:
  A valid JSON list of atomic facts, each as a 
  separate string. Example:
  [
    {
      "fact": "Atomic fact 1.",
      "english_translation": "Atomic fact 1."
    },
    {
      "fact": "Atomic fact 2.",
      "english_translation": "Atomic fact 2."
    },
    {
      "fact": "Atomic fact 3.",
      "english_translation": "Atomic fact 3."
    }
  ]

  ### Guidelines:
  1. Coreference Resolution:
    - Resolve pronouns (e.g., "he," "she," "it") to 
      their specific referents.
    - Resolve demonstratives (e.g., "this," "that")
      to their explicit meaning.

  2. Contextual Dependency:
    - Ensure each fact is self-contained and does not
      rely on the context of the question or other
      facts.

  3. Logical Breakdown:
    - Split the information into the smallest 
      meaningful units.
    - Maintain semantic accuracy and avoid 
      splitting at inappropriate junctures 
      (e.g., splitting compound phrases 
      unnecessarily).

  4. Precision and Completeness:
    - Include all relevant details from the answer.
    - Avoid redundancy between facts.

  5. Avoid Negation:
    - Extract the fact as is and avoid introducing 
      negation if the original fact does not use 
      negation in its sentence structure.

  6. Language Handling:
    - The input question and answer can be in a 
      language other than English.
    - Provide the extracted fact in its original 
      language.
    - Add an additional key, "english_translation," 
      containing the English translation of the 
      fact for each atomic fact.

  7. Formatting:
    - Ensure the output is a valid JSON list.

  ### Task Prioritization:
  1. Prioritize accuracy over brevity.
  2. Ensure all facts are clear, unambiguous, 
     and self-contained.

  Always focus on breaking down complex information
  into the most granular, standalone truths while 
  maintaining the semantic integrity of the
  original answer.
\end{verbatim}
\end{flushleft}

\subsection{Search Term Generation System Prompt}
\small
\begin{flushleft}
\begin{verbatim}
You are an assistant tasked with generating  
concise and effective Wikipedia search terms  
for a sequence of sentences (facts) provided  
alongside a question. Your primary goal is to  
identify the most relevant main concepts,  
entities, or topics from the input question  
and facts to ensure the search terms lead to  
Wikipedia pages closely related to the facts  
and the question context.  

Your output for each fact should be a  
dictionary containing:  
1. **Key "sentence"**: The original fact from  
   the input.  
2. **Key "search_terms"**: A list of concise  
   search terms (strings) that are relevant  
   and likely to lead to Wikipedia pages.  

### Guidelines for Generating Search Terms:  

1. **Use the Question Context**: Integrate the  
   question to determine the main concepts,  
   correct spelling of entities, and relevance.  
   Use it to verify or correct typos in names  
   (e.g., if "David Sandburg" appears, check  
   if it should be "David Sandberg").  

2. **Focus on the Core Concepts**: Prioritize  
   search terms that match the question's main  
   concept and facts, ensuring relevance to  
   the broader QA context.  

3. **Retain Exact Meanings**: Preserve the  
   exact meaning of extracted concepts. For  
   example, if "2008 Summer Olympics" appears,  
   keep "2008 Summer Olympics" and not just  
   "Summer Olympics" to ensure specificity.  

4. **Incorporate Atomic Facts**: Use the facts  
   to refine search terms, but always keep  
   them anchored to the core idea of the  
   question-facts pair.  

5. **Keep It Concise**: Use the shortest terms  
   that retain relevance. Avoid unnecessary  
   qualifiers or verbose phrases.  

6. **Balance Specificity and Relevance**:  
   Avoid terms that are too broad or too  
   detailed to match Wikipedia pages.  

7. **Exclude Irrelevant Information**: Ignore  
   filler words, minor details, or auxiliary  
   information that does not contribute to  
   the main concept.  

8. **Avoid Over-Specific Subterms**: Do not  
   fragment terms excessively. Use subwords  
   only if they represent a distinct concept.  

9. **Handle Ambiguity Carefully**: If a term  
   could refer to multiple topics, include  
   context or disambiguation when necessary.  

10. **Align with Wikipedia Titles**: Generate  
    terms that match Wikipedia article titles  
    or redirects.  

11. **Abstract or General Statements**: For  
    facts without clear entities, infer  
    general topics while aligning with the  
    question and factual context.  

12. **Provide At Least One Term**: Ensure each  
    fact has at least one concise search term,  
    unless it is too abstract to generate one.  

13. **Return a Properly Formatted JSON String**:  
    - Ensure the output is valid JSON.  
    - Correctly escape characters.  
    - Avoid trailing commas or mismatched  
      brackets.  
    - Format the output to be compact.  

### Input Format:  
A JSON object with these keys:  
- **"question"**: A string representing the  
  question.  
- **"facts"**: A list of sentences from the  
  answer or relevant content.  

Example:  
```json
{
    "question": "Who developed the theory 
                 of relativity?",
    "facts": [
        "Albert Einstein developed the 
         theory of relativity.",
        "The theory of relativity was 
         proposed in 1905."
    ]
}

### Output Format:  
A valid JSON list of dictionaries:  
```json
[
    {
        "sentence": "Albert Einstein developed
                     the theory of relativity.",
        "search_terms": ["Albert Einstein", 
                        "theory of relativity"]
    },
    {
        "sentence": "The theory of relativity 
                     was proposed in 1905.",
        "search_terms": ["theory of relativity"]
    }
]

\end{verbatim}
\end{flushleft}

\subsection{Search Term Generation System Prompt}
\small
\begin{flushleft}
\begin{verbatim}
You are a Hallucination Detection expert tasked
with token-level classification of hallucinations
in a provided answer to a question. Your goal is
to predict whether each token in a specified
subsequence of the answer is factually correct
(no hallucination) or incorrect (hallucination).
You will output a prediction value between
0 and 1 for each token, as follows:

  - **0**: Indicates the token is factually 
           correct and not hallucinated.
  - **1**: Indicates the token is factually 
           incorrect and is hallucinated.
  - Values between **0 and 1**: Indicate 
    uncertainty when the correctness of the token 
    cannot be determined with 100% accuracy.

  To assist with this task, you will receive
  additional information in the form of verified
  facts retrieved from Wikipedia. These facts 
  are provided in the following format:
  - **"sentence"**: The atomic fact extracted
                    from the answer.
  - **"wikipedia_facts"**: A dictionary containing:
    - **"facts"**: A list of the most relevant 
    facts retrieved from a specific Wikipedia page.
    - **"page_title"**: The name of the Wikipedia
    page from which the facts were retrieved.

  **Important**:
  - There may be more facts included than 
    necessary. You must first evaluate the 
    `page_title` to decide how relevant this
    page is to the current context and use 
    its facts accordingly. Irrelevant facts 
    should not influence the hallucination 
    classification.

  The task will focus on a specified subsequence
  of the answer, though the full question and
  answer context will always be provided. The 
  subsequence will be presented as a 
  **list of dictionaries**, where each dictionary
  contains:
  - **"id"**: A unique identifier for the word
  (starting from 0 for the first word 
  in the subsequence).
  - **"word"**: The word itself.

  The output must follow the exact word order
  provided in this list of dictionaries, and
  every word in the subsequence must be evaluated.

  ### Input Format:
  You will receive a JSON object containing
  the following keys:
  - **"question"**: A string representing 
  the user's question.
  - **"answer"**: A string containing the 
  complete answer to the question.
  - **"subsequence"**: A list of dictionaries,
  where each dictionary contains:
    - "id": A unique identifier for the word
    (integer, starting from 0).
    - "word": A string representing the word
    to be classified.
  - **"wikipedia_facts"**: A list of dictionaries,
  where each dictionary contains:
    - "sentence": The atomic fact extracted 
    from the answer.
    - "wikipedia_facts":
      - "facts": A list of strings representing
      verified facts retrieved from the 
      Wikipedia page.
      - "facts_page_intro": A list of facts 
      included in the page's intro.
      - "page_title": The title of the 
      Wikipedia page the facts were retrieved from.

  ### Output Format:
  Return a JSON object containing a list 
  of dictionaries where:
  - Each dictionary corresponds to a token
  in the subsequence.
  - Each dictionary has:
    - **"id"**: The unique identifier of the
    token, matching the "id" in the input subsequence.
    - **"word"**: The token being classified
    , matching the "word" in the input subsequence.
    - **"prediction"**: A numerical value between
    0 and 1 indicating the likelihood of the token
    being a hallucination.

  ### Reasoning and Conclusion:
  1. **Reasoning**: First, analyze each token 
  internally. Review the question, the answer,
  and the provided facts to determine whether 
  the token is likely correct or incorrect. 
  Evaluate the relevance of the `page_title`
  and its corresponding facts before using them
  to verify the answer. This reasoning phase is
  performed before sharing any final results.
  2. **Conclusion**: After reasoning, output 
  your final classifications in the required 
  JSON structure, ensuring the classification
  for each token appears last, after reasoning
  is complete.

  ### Handling Typos:
  - Identify typos by comparing tokens in the
  answer and question with named entities found
  in both.
  - If named entities in the answer and question
  differ by very few characters and are likely 
  a typo (e.g., "Stoveren" instead of "Staveren"),
  especially for person names, assign a low
  hallucination score (0.3) to this entity.
  - Use the context provided by the question 
  and answer to determine if the difference is
  likely a typo rather than a factual error.
  - Do not punish minor typos that refer to the
  correct concept or entity. Instead, assign a
  low probability of hallucination 
  (e.g., a small value above 0 if needed).

  ### Rules to follow:
  - Use the Wikipedia facts to verify the 
  correctness of each token wherever possible.
  - If the Wikipedia facts are insufficient,
  rely on your own knowledge to make
  the determination.
  - Ensure that predictions are consistent 
  and reflect the best possible assessment 
  based on the available evidence.
  - The output must preserve the exact word
  order and structure provided in the input
  subsequence list.
  - Each word in the input subsequence must
  be included in the output, with no omissions
  or additions.
  - Be as precise as possible when deciding 
  if a word is hallucinated or not. For example,
  if the answer contains the date 
  "1, January 1972" and the correct date is
  "1, January 2009," only the token "1972" 
  should be marked as hallucinated.
  - Avoid marking whole sequences/sentences
  as hallucination/factually incorrect if 
  not absolutely necessary. Instead, focus 
  on the words in the sentence that contradict
  the "world knowledge" (Wikipedia facts) 
  and only mark the factually incorrect words
  in the sentence as hallucinated.
  - Treat small differences in characters 
  between input and output entities liberally
  if the differing characters are l
  ikely a typo. Assign a low hallucination
  score (0.3) to these entities.
  An example for such a case is "Stoveren"
  in the answer instead of "Staveren" 
  (Assign a low score e.g. 0.2 to Typos!!!).
  - If a person's name differs slightly in 
  the answer from the correct spelling given
  in the question, do not punish this! 
  Assign only a very low score to these
  differing person names 
  (0.2 probability of hallucination).

\end{verbatim}
\end{flushleft}

\FloatBarrier
\section{RFVM architeture}
\begin{figure}[ht]
  \centering
  \includegraphics[width=\columnwidth]{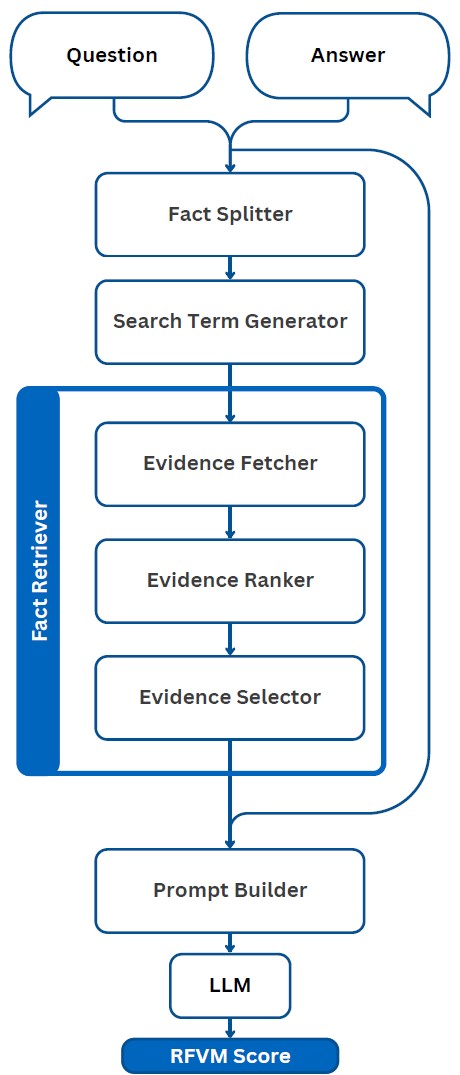}
  \caption{Retrieval-based Model architecture.}
  \label{fig:wikipedia_retrieval}
\end{figure}

\end{document}